%% file: main.tex
\documentclass{article}
\usepackage{arxiv}
\usepackage{graphicx}
\usepackage[numbers]{natbib}
\usepackage{doi}
\usepackage[utf8]{inputenc} 
\usepackage[T1]{fontenc}    
\usepackage{hyperref}       
\usepackage{url}            
\usepackage{booktabs}       
\usepackage{amsfonts}       
\usepackage{nicefrac}       
\usepackage{microtype}      
\usepackage[table]{xcolor}         
\usepackage{amsmath}
\usepackage{amssymb}
\usepackage{array}
\usepackage{pifont}
\usepackage{threeparttable}
\usepackage{graphicx} 
\usepackage{array} 
\usepackage{makecell}
\usepackage{algpseudocode} 
\usepackage{multirow}
\usepackage{listings}
\usepackage{enumitem}
\usepackage{subfig}
\usepackage{hyperref}
\usepackage{todonotes}
\usepackage{subcaption}
\usepackage{float}
\usepackage[ruled,vlined]{algorithm2e}
\usepackage{natbib}
\setcitestyle{numbers,square}

\newcommand{\eg}{\textit{e.g.}}

\newcommand{\ie}{\emph{i.e.}}

\usepackage{authblk}

\newcommand{\pangu}{Pangu}

\usepackage{authblk}

\newcommand{\modelname}{Pangu Light}

\author{\pangu~Team, Huawei\\
pangutech@huawei.com}

\renewcommand{\headeright}{Technical Report}

\title{\modelname: 
Weight Re-Initialization for Pruning and Accelerating LLMs}

\begin{document}

\maketitle

\thispagestyle{fancy}

\begin{abstract}
Large Language Models (LLMs) deliver state-of-the-art capabilities across numerous tasks, but their immense size and inference costs pose significant computational challenges for practical deployment. While structured pruning offers a promising avenue for model compression, existing methods often struggle with the detrimental effects of aggressive, simultaneous width and depth reductions, leading to substantial performance degradation. This paper argues that a critical, often overlooked, aspect in making such aggressive joint pruning viable is the strategic re-initialization and adjustment of remaining weights to improve the model post-pruning training accuracies. We introduce \emph{\modelname}, a framework for LLM acceleration centered around structured pruning coupled with novel weight re-initialization techniques designed to address this ``missing piece''. Our framework systematically targets multiple axes, including model width, depth, attention heads, and RMSNorm, with its effectiveness rooted in novel re-initialization methods like Cross-Layer Attention Pruning (CLAP) and Stabilized LayerNorm Pruning (SLNP) that mitigate performance drops by providing the network a better training starting point. Further enhancing efficiency, \modelname~incorporates specialized optimizations such as absorbing Post-RMSNorm computations and tailors its strategies to Ascend NPU characteristics. The \modelname~models consistently exhibit a superior accuracy-efficiency trade-off, outperforming prominent baseline pruning methods like Nemotron and established LLMs like Qwen3 series. For instance, on Ascend NPUs, \modelname-32B's 81.6 average score and 2585 tokens/s throughput exceed Qwen3-32B's 80.9 average score and 2225 tokens/s.
\end{abstract}

\section{Introduction}

Large Language Models (LLMs) have revolutionized natural language processing, achieving remarkable proficiency across diverse and complex tasks \cite{guo2025deepseek,dubey2024llama,ai2023gpt}. Models such as the Pangu series~\cite{Yin2025PanguUP} stand at the forefront of these advancements. However, the immense scale required to achieve state-of-the-art performance leads to substantial computational costs, particularly during inference, creating barriers to widespread adoption and deployment in practical applications.

Model compression techniques aim to alleviate this burden. Structured pruning \cite{frantar2023sparsegpt,ma2023llm,xia2023sheared,men2024shortgpt,pei2024fusegpt}, which removes structural components like neurons, attention heads, or layers, and knowledge distillation (KD) \cite{hinton2015distilling}, where a smaller model learns from a larger one, are widely explored avenues. Recent studies like Minitron \cite{muralidharan2024compact} have combined pruning across various axes incluidng depth, width, and attention with KD-based retraining. Frameworks like PUZZLE \cite{bercovich2024puzzle} employ Neural Architecture Search (NAS) with distillation for hardware-specific optimizations.

Despite these advances, a critical challenge persists: aggressive, simultaneous pruning of model width and depth often leads to a drastic collapse in performance. While current strategies excel at identifying structural components like layers, attention heads, or neurons for removal based on various importance metrics, they often fail to address the profound instability and representational damage inflicted by such extensive alterations. The model's internal dynamics can be so heavily perturbed that even extensive retraining or knowledge distillation struggles to fully recover. We posit that a crucial, yet largely underexplored, element for successful aggressive joint pruning is a dedicated phase of weight re-initialization or targeted adjustment immediately following structural removal. This step is vital to pull the model back from a severely degraded state and provide a more stable foundation for subsequent recovery or fine-tuning, effectively acting as a ``missing piece'' in current joint pruning methodologies. Moreover, fine-grained optimizations for components like normalization layers, which contribute to inference latency especially on specialized hardware such as Ascend NPUs where vector-bound operations are relatively costly~\cite{Yin2025PanguUP,lin2024fastattention}, remain less explored within comprehensive pruning frameworks.

In this paper, we introduce \emph{\modelname}, a novel and comprehensive framework designed to address these challenges and enable effective acceleration of LLMs by explicitly incorporating this ``missing piece''. \modelname~builds upon established importance-driven structured pruning for initial component removal across model width, depth, and attention heads. Its core innovation lies in the introduction of sophisticated weight re-initialization strategies, notably Cross-Layer Attention Pruning (CLAP) and Stabilized LayerNorm Pruning (SLNP). These techniques are specifically designed to counteract the detrimental effects of joint pruning by restabilizing the pruned network and preserving crucial information, thereby facilitating robust performance recovery. Furthermore, leveraging insights into the Pangu model architecture and its prevalent Sandwich-Norm structure, \modelname~incorporates a specialized Post-RMSNorm absorption technique that converts costly normalization steps into efficient, fusable operations, further enhancing inference speed. This unified approach synergistically combines principled pruning, novel weight re-initialization mechanisms, targeted normalization optimization. We provide conceptual illustrations for our weight re-initialization strategies and normalization optimization in Figure 1.

\begin{figure}[tbp!]
\centering
\includegraphics[width=\textwidth]{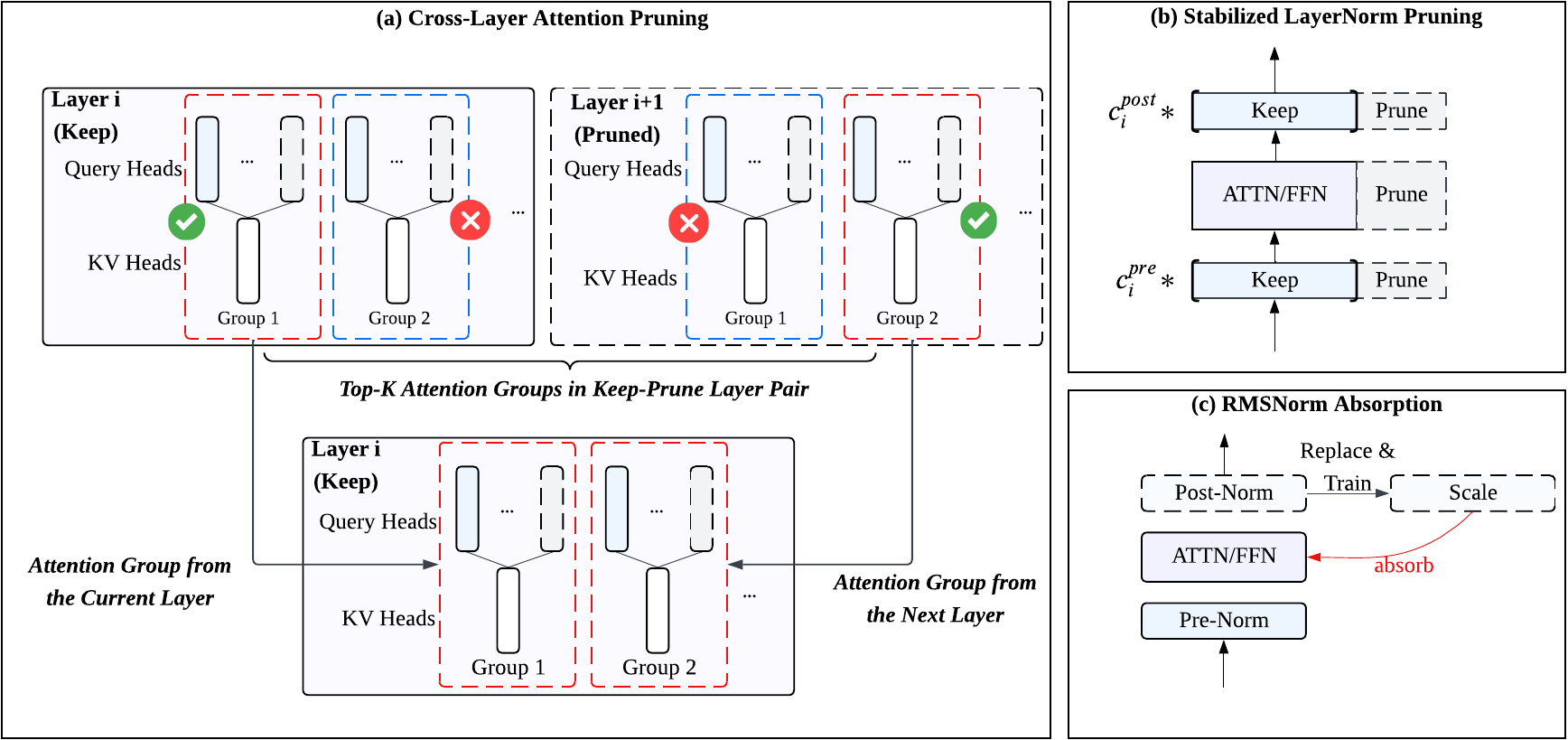} 
\caption{Conceptual overview of the \modelname~methodology, illustrating its integrated approach that combines importance-based structural pruning with novel weight re-initialization strategies (a)(b) and specialized normalization layer optimization (c).}
\label{fig:pruning}
\end{figure}
To validate the practical effectiveness of the \modelname~methodology, we applied it to compress the Pangu-38B model, yielding the \modelname~series. Evaluated on Ascend NPUs, these models substantiate our claims. For example, \modelname-32B not only outperforms comparable models like Qwen3-32B in reasoning tasks, achieving an average score of 81.6 to Qwen3-32B's 80.9, but also delivers superior throughput, registering 2585 tokens/s versus Qwen3-32B's 2225 tokens/s. Furthermore, this approach, integrating targeted weight re-initialization, enables aggressive pruning with minimal performance degradation and offers a compelling alternative to existing methods. It achieves, for instance, comparable acceleration to PUZZLE~\cite{bercovich2024puzzle}: \modelname~provides approximately 2.1x speedup while retaining 98.9\% accuracy, which is higher than PUZZLE's reported 98.4\% retention for similar acceleration levels. These results highlight a clear advantage in the accuracy-efficiency trade-off, underscoring the efficacy of our comprehensive strategy.

\input{sec/related_work}
\input{sec/pangu-f}
\input{sec/exp}
\section{Conclusion}
In this paper, we introduced \emph{\modelname}, a comprehensive framework designed to address critical challenges in LLM acceleration. \modelname~advances the state-of-the-art through a systematical structured pruning strategy that uniquely integrates principled component removal with sophisticated parameter and structural weight re-initilization techniques. Our core innovations, Cross-Layer Attention Pruning (CLAP) and Stabilized LayerNorm Pruning (SLNP), were specifically developed to counteract the detrimental effects of joint pruning by preserving crucial information and stabilizing model outputs. Furthermore, \modelname~incorporates a specialized Post-RMSNorm absorption technique, tailored for architectures like Pangu's Sandwich-Norm, to convert costly normalization steps into efficient, fusable operations, thereby enhancing inference speed without compromising model integrity. This entire framework, complemented by knowledge distillation during performance recovery, was carefully co-designed considering the architectural characteristics of Ascend NPUs to maximize efficiency gains.

The \modelname~framework represents a significant step towards making state-of-the-art LLMs more efficient and accessible. By systematically addressing not only what to prune but critically how to compensate for aggressive structural modifications and optimize often-overlooked components, our work provides a robust and effective methodology for practical LLM deployment. Future work may explore the extension of these principles to other model architectures and hardware platforms, as well as the integration of \modelname~with other compression techniques like quantization for even greater efficiency.

\bibliographystyle{plain}
\bibliography{ref}

\newpage
\appendix

\section{Contributions and Acknowledgments}

\noindent
\textbf{Core Contributors}
Hanting Chen, Jiarui Qin, Jialong Guo, Tao Yuan, Yichun Yin, Huiling Zhen, Yasheng Wang, Jinpeng Li, Xiaojun Meng, Meng Zhang, Rongju Ruan, Zheyuan Bai, Yehui Tang, Can Chen, Xinghao Chen, Fisher Yu, Ruiming Tang, Yunhe Wang 

\noindent
\textbf{Contributors}
Bin Wang, Boxiao Liu, Dingyu Yong, Dong Li, Fei Mi, Hui Zang, Jiansheng Wei, Kaikai Song, Liqun Deng, Luocheng Hu, Wenyong Huang, Xianzhi Yu, Xu He, Yixian Ren, Yufei Wang, Yuhang Gai, Zhao Liu, Ziyang Zhang

\end{document}

%% file: sec/related_work.tex
\section{Related Work}
\textbf{Granularity of LLM pruning.} Based on the granularity of model pruning, pruning methods can be divided into structured pruning and unstructured pruning. Unstructured pruning aims to eliminate unimportant connections in the network to construct sparse matrices. There are some related works such as SparseGPT~\cite{frantar2023sparsegpt} and Wanda~\cite{sun2023simple}. To reduce inference latency, unstructured pruning typically requires specific hardware support. Therefore, it is difficult to achieve acceleration on edge devices. Structured pruning typically compresses models by eliminating rows or columns of the weight matrix. Current research primarily focuses on two aspects: width pruning and depth pruning. Width pruning generally adopts the method of compressing the size of the weight matrix. Many works, such as LLMPruner~\cite{ma2023llm}, FLAP~\cite{an2024fluctuation}, and SlimGPT~\cite{ling2024slimgpt}, reduce the model width by decreasing the number of channels and attention heads. In addition, SliceGPT~\cite{ashkboos2024slicegpt} designs strategies to reduce the dimension of the embedding layer, while LoRAP~\cite{li2024lorap} proposes the AWSVD method to compress the parameter matrices of MHA. Depth pruning takes layers as the pruning units. Related works include ShortGPT~\cite{men2024shortgpt}, Laco~\cite{yang2024laco}, and Shorten LLaMa~\cite{kim2024shortened}. Meanwhile, several works apply both depth and width pruning, like Minitron~\cite{muralidharan2024compact}.

\textbf{Metric for LLM pruning.} For LLM pruning, various criteria for determining importance have emerged recently. Some works, like Sheared LLaMA~\cite{xia2023sheared}, introduce learnable parameters to compress LLMs into target structures. However, these methods typically require a sufficient amount of data. To reduce the cost of compression, some efficient pruning methods have been proposed. For width pruning, the main strategies for determining importance can be divided into gradient-based methods and activation/weight-based methods. Gradient-based methods~\cite{ma2023llm} require more computation and memory. To address this issue, LORAPrune~\cite{zhang2023loraprune} employs a LORA~\cite{hu2021lora} training strategy to reduce computational cost. In contrast, activation/weight-based strategies can determine the importance of pruning units based solely on forward computation. Among them, Minitron~\cite{muralidharan2024compact} only uses activation to determine the importance of channels or heads. FLAP~\cite{an2024fluctuation} combines the variance of activations and weights to design a fluctuation metric. LORAP~\cite{li2024lorap} and Wanda~\cite{sun2023simple} use the magnitude of activations and weights as criteria for importance. SlimGPT~\cite{ling2024slimgpt} applies the Optimal Brain Surgeon framework to structured pruning. In terms of depth pruning, recent methods have largely employed strategies based on cosine similarity~\cite{men2024shortgpt} or PPL~\cite{kim2024shortened} to determine the importance of each layer.

\textbf{Normalization for LLM.} Normalization is an essential component of current models. Among various types, LayerNorm (LN)~\cite{ba2016layer} is a commonly used normalization structure in Transformer models. Subsequently, RMSNorm~\cite{zhang2019root} simplifies the LN strategy and is adopted by many LLMs. In a recent work, Dynamic Tanh~\cite{zhu2025transformers} proposes a strategy for replacing normalization layers, which achieves performance comparable to LN with higher computational efficiency. Given that LN accounts for a significant proportion of inference latency in LLMs, we believe that finding an effective replacement strategy for LN is highly meaningful.

%% file: sec/pangu-f.tex
\section{Methodology}
This section elaborates on our comprehensive pruning strategy, designed for a full-axis reduction of Pangu LLMs, encompassing aspects such as layers, network width, and attention heads. We begin by detailing our methodology for assessing the importance of each structural component (Section \ref{sec:importance_metrics}), which informs what to prune. Subsequently, we introduce our novel {weight re-initialization and structural adjustment strategies} (Section \ref{sec:pruning_stabilization}), which we argue are critical for stabilizing the model and recovering performance after aggressive joint pruning across multiple axes---addressing the ``missing piece'' in such scenarios. We then present a specialized normalization layer optimization tailored to the Sandwich-Norm architecture in Pangu LLMs (Section \ref{sec:sandwich_absorption_method}), followed by our approach to performance recovery using knowledge distillation (Section \ref{sec:kd_recovery}).

\subsection{Multi-Axis Importance Metrics}
\label{sec:importance_metrics}
Multi-axis pruning approaches \cite{muralidharan2024compact, bercovich2024puzzle} often yield superior results compared to single-axis pruning, primarily due to the increased flexibility in the resulting pruned architectures. Therefore, we adopt an importance measurement strategy largely inspired by \cite{muralidharan2024compact}. Our network pruning strategy evaluates the importance of hidden channels, attention heads, feed-forward network (FFN) neurons, and entire layers using activation statistics from a calibration dataset $\mathcal{C}$. This dataset comprises a collection of input sequences. Based on calculated importance scores, components are pruned if their scores fall below a threshold or rank.

\paragraph{Hidden Channel Importance}
The significance of each hidden channel $k$, where $k \in \{1, \dots, d\}$ and $d$ is the hidden dimensionality, is determined by aggregating the L2-norm of its activation values from all RMSNorm layers across the model. The global importance score for channel $k$, $S_{\text{channel}}^{k}$, is calculated as
\begin{equation} \label{eq:channel_importance}
S_{\text{channel}}^{k} = \sum_{l=1}^{L}  \sum_{\mathbf{X} \in \mathcal{X}_l} \vert\vert \text{RMSNorm}_l(\mathbf{X})_k \vert\vert_2,
\end{equation}
where $L$ is the total number of RMSNorm layers in the model; $\mathcal{X}_l$ is the multiset of $d$-dimensional token representations $\mathbf{X}$ that are input to the RMSNorm component of layer $l$, derived from the calibration set $\mathcal{C}$; $\text{RMSNorm}_l(\mathbf{X})$ is the $d$-dimensional output vector from the RMSNorm unit of layer $l$ for an input token representation $\mathbf{X}$; and $\text{RMSNorm}_l(\mathbf{X})_k$ is the scalar value of the $k$-th channel of this output. This global score $S_{\text{channel}}^{k}$ guides the selection of top-K channels uniformly across all layers, ensuring consistent hidden representation dimensions. Consequently, the embedding table, the final output layer, and the affine weight parameters $\boldsymbol{\gamma}$ of all RMSNorm layers are pruned by removing dimensions corresponding to channels with low $S_{\text{channel}}^{k}$ scores.

\paragraph{Attention Head Importance}
The importance of each attention head $j$ within a specific layer $l$ is computed based on the L2-norm of its output activations, with the score $S_{\text{head},l}^{j}$ given by
\begin{equation} \label{eq:head_importance}
S_{\text{head},l}^{j} = \sum_{\mathbf{X} \in \mathcal{X}_l} \left\| \text{AttnHead}_j(\mathbf{X}; \mathbf{W}_{l,j}^Q, \mathbf{W}_{l,j}^K, \mathbf{W}_{l,j}^V) \right\|_2,
\end{equation}
where $\mathcal{X}_l$ here refers to the multiset of token representations $\mathbf{X}$ input to the attention mechanism in layer $l$; $\mathbf{W}_{l,j}^Q \in \mathbb{R}^{d \times d_{\text{head}}}$, $\mathbf{W}_{l,j}^K \in \mathbb{R}^{d \times d_{\text{head}}}$, and $\mathbf{W}_{l,j}^V \in \mathbb{R}^{d \times d_{\text{head}}}$ are the query, key, and value projection matrices for attention head $j$ in layer $l$ ($d_{\text{head}}$ being the head dimension); and $\text{AttnHead}_j(\cdot)$ is the $d_{\text{head}}$-dimensional output vector of head $j$.
Due to the Pangu models' use of Grouped Query Attention (GQA), heads are pruned within their respective Key-Value (KV) groups. For the query projection matrix $\mathbf{W}_{l}^Q \in \mathbb{R}^{d \times N_H d_{\text{head}}}$, where $N_H$ is the number of query heads in layer $l$), input dimensions are pruned based on $S_{\text{channel}}^{k}$, while output dimensions are pruned based on the selected top-K $S_{\text{head},l}^{j}$ scores within each KV group. For the output projection matrix $\mathbf{W}_{l,O} \in \mathbb{R}^{N_H d_{\text{head}} \times d}$, input dimensions are removed, and output dimensions are pruned according to $S_{\text{channel}}^{k}$. The input dimensions of KV projection matrices $\mathbf{W}_{l}^K$ and $\mathbf{W}_{l}^V$ are pruned by $S_{\text{channel}}^{k}$, but their output dimensions are not pruned in this step to preserve the GQA structure. The importance scores $S_{\text{head},l}^{j}$ are calculated and applied on a per-layer basis.

\paragraph{FFN Activation Importance}
The importance of an intermediate neuron $m$ in the FFN of layer $l$, where $m \in \{1, \dots, d_{\text{ffn}}\}$ and $d_{\text{ffn}}$ is the FFN intermediate dimension, is determined by the magnitude of its activation. The score $S_{\text{activation},l}^{m}$ is calculated as
\begin{equation} \label{eq:ffn_importance}
S_{\text{acticvation},l}^{m}  = \sum_{\mathbf{X} \in \mathcal{X}_l} \left\|\left( \text{Swish}(\mathbf{X}\mathbf{W}_{l,1}^{\text{gate}}) \odot \mathbf{X}\mathbf{W}_{l,1}^{\text{up}} \right)_m \right\|_2,
\end{equation}
where $\mathcal{X}_l$ denotes the multiset of token representations $\mathbf{X}$ input to the FFN in layer $l$; $\mathbf{W}_{l,1}^{\text{gate}} \in \mathbb{R}^{d \times d_{\text{ffn}}}$ and $\mathbf{W}_{l,1}^{\text{up}} \in \mathbb{R}^{d \times d_{\text{ffn}}}$ are the gate and up projection matrices for the first FFN linear transformation in layer $l$; $\odot$ denotes the element-wise product; and $(\cdot)_m$ selects the $m$-th component from the resulting $d_{\text{ffn}}$-dimensional intermediate activation vector. For the matrices $\mathbf{W}_{l,1}^{\text{gate}}$ and $\mathbf{W}_{l,1}^{\text{up}}$, input dimensions are pruned based on $S_{\text{channel}}^{k}$, while output dimensions are selected using their layer-specific importance $S_{\text{activation},l}^{m}$. For the second FFN matrix $\mathbf{W}_{l,2} \in \mathbb{R}^{d_{\text{ffn}} \times d}$, input dimensions are pruned based on $S_{\text{activation},l}^{m}$, and output dimensions are pruned according to $S_{\text{channel}}^{k}$.

\paragraph{Layer Importance}
To identify and remove less salient layers, we utilize a Block Importance (BI) metric, adapted from \cite{Men2024ShortGPTLI}. The BI score for layer $l$, $S_{\text{layer}}^{l}$, measures the average cosine distance between each input token representation to layer $l$ and its corresponding output token representation. This average is computed over the multiset $\mathcal{X}_l$ of all $d$-dimensional token representations $\mathbf{X}$ input to layer $l$ from the calibration set $\mathcal{C}$ by
\begin{equation} \label{eq:layer_importance}
S_{\text{layer}}^{l} = 1 - \mathbb{E}_{\mathbf{X} \in \mathcal{X}_l} \left[ \frac{\mathbf{X} \cdot \text{Layer}_l(\mathbf{X})}{\| \mathbf{X} \|_2 \| \text{Layer}_l(\mathbf{X}) \|_2} \right],
\end{equation}
where $\mathbf{X} \in \mathcal{X}_l$ is an input token representation to layer $l$, and $\text{Layer}_l(\mathbf{X})$ is the corresponding $d$-dimensional output token representation after transformation by the entirety of layer $l$. The expectation $\mathbb{E}_{\mathbf{X} \in \mathcal{X}_l}[\cdot]$ denotes the average taken over all token representations in $\mathcal{X}_l$. A larger cosine distance, which denotes higher $S_{\text{layer}}^{l}$ score, signifies that layer $l$ substantially transforms its input tokens, indicating greater importance. Conversely, layers with lower BI scores are considered less critical and become candidates for removal.

\subsection{Strategic Weight Re-initialization and Structural Adjustment for Pruned Models}
\label{sec:pruning_stabilization}
While the importance-based pruning described before effectively identifies redundant parameters, the aggressive, multi-axis nature of this pruning, which is simultaneously reducing layers and network channels, can render the model architecture highly sensitive. Direct application of such joint pruning often leads to a substantial degradation in model accuracy, as the model's learned weights and internal dynamics are severely perturbed. To address this critical issue---the ``missing piece'' in aggressive pruning---we introduce dedicated strategies focused on {weight re-initialization and structural adjustment}. These are tailored to both width and depth reductions to re-stabilize the model and enable it to retain a high degree of its original performance, forming a crucial step before any subsequent fine-tuning.

\subsubsection{Depth Stabilization: Cross-Layer Attention Pruning (CLAP) for Parameter Re-integration}
\label{sec:clap}
Aggressively pruning entire layers can lead to significant information loss and network instability, as crucial computational pathways are severed. We hypothesize that vital attention-based computations are often distributed or refined across consecutive layers. Simply removing a layer might discard unique, valuable features not entirely redundant with those in the preceding layer. To counteract the severe impact of layer removal, we propose Cross-Layer Attention Pruning (CLAP), a method involving {strategic parameter re-integration and re-initialization}. CLAP aims to preserve and transfer critical attention capabilities from a pruned layer by selectively merging its most important Key-Value (KV) groups' parameters into the preceding layer. This carefully re-initializes and augments the recipient layer's attention mechanism, rather than simply discarding the pruned layer's contributions. This concept of preserving information from pruned layers has shown benefits for performance \cite{Chen2021bert2BERTTR}.

The CLAP mechanism operates as follows: for any pair of consecutive layers, $l$ and $l+1$, if layer $l+1$ is designated for pruning, its attention apparatus is not entirely discarded. Instead, its Key-Value (KV) groups are considered for integration into layer $l$. The process, illustrated in Figure~\ref{fig:pruning}(a), involves:
\begin{enumerate}
    \item \textbf{Initial Head Pruning:} Standard head pruning is first applied independently to the query heads within each KV group of layer $l$ and layer $l+1$.
    \item \textbf{KV Group Importance Scoring:} After initial head pruning, we calculate an importance score for each KV group. For a KV group $G_g$, its score $S_{\text{kv-group}}^{g}$ is defined as the average importance of its constituent query heads that survived the initial head pruning:
    \begin{equation} \label{eq:kv_group_importance}
    S_{\text{kv-group}}^{g} = \frac{1}{N_q(G_g)} \sum_{j \in \text{Heads}_{\text{s}}(G_g)} S_{\text{head}, \text{ori}(j)}^{j},
    \end{equation}
    where $\text{Heads}_{\text{s}}(G_g)$ is the set of query heads remaining in KV group $G_g$ after the initial head pruning, $N_q(G_g)$ is the number of such heads, and $S_{\text{head}, \text{ori}(j)}^{j}$ is the previously computed importance score for query head $j$ from its original layer.
    \item \textbf{Joint Group Ranking and Selection:} All KV groups from layer $l$ and the to-be-pruned layer $l+1$ are ranked together based on their $S_{\text{kv-group}}^{g}$ scores.
    \item \textbf{Parameter Merging and Re-initialization:} The top-K ranked KV groups from this combined list are selected to constitute the attention mechanism of the modified layer $l$. If a selected KV group originated from layer $l+1$, its associated parameters, \ie, the K and V projection parameters for that group, and the Q projection and output projection parameters corresponding to its surviving query heads, are transferred and appropriately integrated into layer $l$, effectively {re-initializing} parts of layer $l$'s attention mechanism with high-importance parameters from the pruned layer $l+1$. This transfer is performed group-wise to maintain the integrity of query-KV mappings.
\end{enumerate}
We also explored applying a similar strategy of parameter merging for FFN blocks, but found it did not yield significant performance improvements in our experiments; thus, our primary focus for structural re-adjustment post-depth-pruning is on the attention mechanisms via CLAP.

\subsubsection{Width Stabilization: Stabilized LayerNorm Pruning (SLNP) for Weight Re-initialization}
When hidden channels are pruned, the affine parameters $\boldsymbol{\gamma}$ of the RMSNorm layers are also reduced in dimension. This abrupt change can drastically alter the L2-norm of the RMSNorm output, leading to activation instability in subsequent layers and hindering convergence during post-pruning fine-tuning. We observed that re-establishing the output scale of these normalization layers via {direct weight re-initialization} is crucial for model stability.
To address this, we introduce Stabilized LayerNorm Pruning (SLNP), a {weight re-initialization technique} for the $\boldsymbol{\gamma}$ parameters of RMSNorm layers. For each RMSNorm layer $l$ whose $\boldsymbol{\gamma}_l$ parameters are pruned, we compute a {re-initialization} scalar
\begin{equation} \label{eq:ln-compensation}
c_l = \frac{\|\boldsymbol{\gamma}_l^{\text{orig}}\|_2}{\|\boldsymbol{\gamma}_l^{\text{pruned}}\|_2},
\end{equation}
where $\boldsymbol{\gamma}_l^{\text{orig}}$ represents the original affine parameters of the $l$-th RMSNorm layer before pruning, and $\boldsymbol{\gamma}_l^{\text{pruned}}$ are the corresponding parameters after channel pruning. This scalar $c_l$ is then used to rescale the pruned $\boldsymbol{\gamma}_l^{\text{pruned}}$ parameters ,\ie, the re-initialized parameters $\boldsymbol{\gamma}_l^{\text{new}} = c_l \times \boldsymbol{\gamma}_l^{\text{pruned}}$, effectively restoring the overall magnitude of the affine transformation through targeted {re-initialization}. We found this {re-initialization step} particularly important for stabilizing the model before subsequent fine-tuning or continued training phases.

\subsection{Optimization for Sandwich-Norm Architectures via Parameter Absorption}
\label{sec:sandwich_absorption_method}
This subsection details a specialized approach for handling the unique Sandwich-Norm structure present in Pangu LLMs, aiming to reduce its inference overhead by effectively eliminating Post-RMSNorm layers while preserving model integrity {through parameter absorption and fusion}.

\paragraph{The Pangu Depth-Scaled Sandwich-Norm (DSSN) Architecture}
Pangu LLMs employ a distinctive Depth-Scaled Sandwich-Norm (DSSN) architecture. A hallmark of this design is the strategic placement of an additional RMSNorm layer after both the attention and the FFN modules within each Transformer block \cite{Yin2025PanguUP}. This ``sandwich'' configuration, with normalization layers bracketing the main computational modules, significantly enhances training stability and has been instrumental in achieving pre-training trajectories with {zero loss spikes}.

However, while beneficial for training, the Sandwich-Norm structure, particularly the post-module RMSNorm layers, \ie, Post-RMSNorm, introduces notable computational overhead during inference due to the extra normalization operations. For instance, in a 38-billion parameter Pangu model, eliminating these Post-RMSNorm layers can yield a throughput enhancement of up to 6\% on Ascend NPUs. Furthermore, these Post-RMSNorm layers require careful consideration during hidden-size pruning, as their learnable affine parameters ($\boldsymbol{\gamma}$) directly modulate the output magnitudes of the attention and FFN modules, influencing the overall network dynamics.

\paragraph{Absorbing Post-RMSNorm for Inference Acceleration}
The enhanced stability offered by the Post-RMSNorm layers is particularly crucial during the initial and intermediate stages of large model training. However, we observe that towards the end of training, as the model converges, the variance of activations entering these Post-RMSNorm layers tends to stabilize. This suggests that the dynamic normalization provided by these layers becomes less critical, and their computational effect can be approximated or absorbed into preceding layer parameters to improve inference efficiency.

The standard RMSNorm operation is typically applied to each  token vector $\mathbf{x} \in \mathbb{R}^d$ within an input tensor, \eg, $\mathbf{X} \in \mathbb{R}^{S \times d}$ representing a sequence of $S$ tokens. With learnable affine scale parameters $\boldsymbol{\gamma} \in \mathbb{R}^d$, it is defined as
\begin{equation} \label{eq:rmsnorm_definition}
\text{RMSNorm}(\mathbf{x}) = \boldsymbol{\gamma} \odot \frac{\mathbf{x}}{\sqrt{\frac{1}{d}\|\mathbf{x}\|_2^2 + \epsilon}},
\end{equation}
where $\|\mathbf{x}\|_2^2 = \sum_{j=1}^{d} x_j^2$ is the squared L2-norm of the token vector $\mathbf{x}$ with $x_j$ being its $j$-th component, $\epsilon$ is a small constant, \eg, $10^{-6}$, for numerical stability, and $\odot$ denotes element-wise multiplication. The Post-RMSNorm layers, by applying this operation to the output of each attention and FFN module, contribute to unavoidable runtime costs during inference.

To mitigate this, we propose to absorb the Post-RMSNorm operation into a static scaling factor by {modifying its affine parameters}. Instead of computing the instance-specific normalization term $1/\sqrt{\frac{1}{d}\|\mathbf{x}\|_2^2 + \epsilon}$ in real-time for each token, we pre-calculate its expected value over a representative calibration multiset $\mathcal{X}_l$. This multiset comprises all individual $d$-dimensional token vectors encountered when processing a broader calibration dataset $\mathcal{C}$. Let this average inverse scaling magnitude be
\begin{equation} \label{eq:avg_inv_scale_factor}
\bar{s}_{\text{inv}} = \mathbb{E}_{\mathbf{x} \sim  \mathcal{X}_l} \left[ \frac{1}{\sqrt{\frac{1}{d}\|\mathbf{x}\|_2^2 + \epsilon}} \right] \approx \frac{1}{| \mathcal{X}_l|} \sum_{\mathbf{x} \in  \mathcal{X}_l} \frac{1}{\sqrt{\frac{1}{d}\|\mathbf{x}\|_2^2 + \epsilon}}.
\end{equation}
The original learnable parameters $\boldsymbol{\gamma}$ of the Post-RMSNorm layer are then {updated by absorbing} this average scaling factor, yielding new effective affine parameters
\begin{equation} \label{eq:gamma_update_absorption}
\boldsymbol{\gamma}_{\text{abs}} = \bar{s}_{\text{inv}} \times \boldsymbol{\gamma}.
\end{equation}
With $\boldsymbol{\gamma}_{\text{abs}}$, the entire Post-RMSNorm operation on an input tensor $\mathbf{x}$, which is the output from a preceding attention or FFN module, is simplified. For each token vector $\mathbf{x}$ in $\mathbf{X}$, the operation becomes an element-wise scaling:
\begin{equation} \label{eq:absorbed_norm_operation}
\text{AbsorbedPostNorm}(\mathbf{x}) = \boldsymbol{\gamma}_{\text{abs}} \odot \mathbf{x}.
\end{equation}
This transformation effectively replaces the Post-RMSNorm layer with a constant channel-wise scaling. Crucially, this scaling operation can be seamlessly fused into the weight matrix of the linear projection layer that produces $\mathbf{X}$, \eg, the output projection of an attention or FFN module, $\mathbf{W}_{\text{proj}}$, where $\mathbf{X} = \text{Input} \times \mathbf{W}_{\text{proj}}$). The absorbed operation $\boldsymbol{\gamma}_{\text{abs}} \odot \mathbf{X}$ is mathematically equivalent to using a modified projection matrix $\mathbf{W}'_{\text{proj}}$ where each column $j$ is scaled by the $j$-th component of $\boldsymbol{\gamma}_{\text{abs}}$, \ie, $(\mathbf{W}'_{\text{proj}})_{:,j} = (\mathbf{W}_{\text{proj}})_{:,j} \times (\boldsymbol{\gamma}_{\text{abs}})_j$. This fusion entirely eliminates the computational overhead of the Post-RMSNorm layer during inference, effectively reducing its operational cost to zero. This accelerates model execution, particularly when applied after model convergence, without significantly impacting the learned representations.

\subsection{Performance Recovery with Knowledge Distillation}
\label{sec:kd_recovery}
Following the structural pruning, strategic weight re-initialization, and architectural optimizations like Post-RMSNorm absorption, the model undergoes a performance recovery phase. For this, we employ knowledge distillation, using the original, unpruned model as the teacher. The distillation is implemented in an online manner, calculating the distillation loss on the full logits across the entire vocabulary. To ensure efficiency, the teacher and student models utilize the same parallel settings during this phase. This phase is crucial for fine-tuning the compressed and re-stabilized model to regain its performance on downstream tasks.

%% file: sec/exp.tex
\section{Experiments}
This section presents a comprehensive evaluation of our proposed pruning methodology, \modelname. We showcase the performance of models pruned using \modelname~across various compression ratios, demonstrating its efficacy in significantly reducing the size of large language models. Furthermore, we conduct detailed ablation studies to dissect the contribution of each component within our pruning strategy. Finally, an analysis of model parameters before and after pruning is provided to offer deeper insights into the structural changes induced by our approach.

\subsection{Experimental Setup}

\paragraph{Pruned Models}
The base model for all our pruning experiments is Pangu-38B. To thoroughly validate our methodology, we apply it to achieve various compression rates, incorporating both layer reduction and hidden dimension/channel reduction. The resultant throughput of the pruned models are detailed in Table~\ref{tab:throughput}. These specific pruned configurations were determined through a co-evaluation process that balanced inference efficiency on Ascend NPUs with preliminary model performance metrics obtained after an initial pruning phase. Specifically, we employed simulation tools to estimate the inference speeds of different architectural configurations under given parameter constraints. Concurrently, we assessed a preliminary loss metric after the initial application of pruning. The final architectures were then selected by optimizing this trade-off between predicted inference speed and early performance indicators.

We evaluated the inference throughput of our pruned \modelname~models against the original Pangu-38B and comparable Qwen models. All performance metrics were benchmarked on a server equipped with Ascend NPUs. The evaluation workload was designed to reflect typical usage patterns, based on daily operational statistics, and comprised an equidistant mix of two distinct settings: a first setting with 128 input tokens and 1024 decoded tokens, and a second setting with 256 input tokens and 2048 decoded tokens.

The detailed throughput results are presented in Table~\ref{tab:throughput}. It is noteworthy that when models are pruned to a comparable 32B scale, our \modelname-32B variant, which is also called \modelname-1.6x, is approximately 16.2\% faster than the Qwen3-32B model. This performance advantage is attributed to several factors, including Qwen3's use of Query-Key (QK) normalization, and the inherent architectural design of \modelname~models which is more effectively optimized for Ascend NPUs.

\begin{table}[htbp]
    \centering
    \caption{Throughput (tokens/s) of different Pangu and Qwen models on Ascend NPUs. \modelname~variants are compared against the original Pangu model and Qwen models of similar sizes.}
    
    \begin{tabular}{c|c|ccc|cc|cc}
        \toprule
         & \multicolumn{1}{c|}{\textbf{Pangu}} & \multicolumn{3}{c|}{\textbf{\modelname}}& \multicolumn{2}{c|}{\textbf{Qwen2.5}} & \multicolumn{2}{c}{\textbf{Qwen3}}\\
        \textbf{Model} & \textbf{38B } & \textbf{1.6x (32B) } & \textbf{2.1x } & \textbf{4.2x } & \textbf{32B} & \textbf{14B} & \textbf{32B} & \textbf{14B}\\
        \midrule
        \textbf{Throughput (tokens/s)} & 1631 & 2585 & 3403 & 6831 & 2316 & 6467 & 2225 & 6254 \\
        \bottomrule
    \end{tabular}
    \label{tab:throughput}
\end{table}

\subsection{Evaluation Setup}

\paragraph{Continue-Training for Performance Recovery}
To regain performance lost during the pruning process, the pruned models undergo a continue-training phase. The pruned models, derived from the original Pangu-38B, are trained on a 300-billion-token dataset specifically curated for this ``annealing'' stage. The learning rate follows a cosine decay schedule, beginning at $1 \times 10^{-5}$ and decreasing to $1 \times 10^{-7}$. During this recovery phase, we employ knowledge distillation, using the original Pangu-38B model as the teacher to provide target logits for the pruned student models. The dataset for this continue-training phase consists of 300 billion tokens, sampled from high-quality sections of our pre-training corpora. This data includes 200 billion tokens formatted with a 32K sequence length and 100 billion tokens with an extended 144K sequence length, facilitating adaptation to long-context capabilities. The data mixture is intentionally diverse, covering a broad spectrum of topics, and is significantly enriched with over 20\% instruction-following data and 36\% reasoning-intensive data, such as mathematics and programming code. To bolster performance on complex reasoning tasks, we have also incorporated carefully designed internal question banks that feature clear and logical chain-of-thought demonstrations. The quality of this training data is continuously monitored and improved using model-based evaluation, leveraging a fine-tuned evaluator, more details of which can be found in~\cite{Yin2025PanguUP}.

\paragraph{Post Training Details}
In the post-training phase, we synthesize a large amount of long CoT data, using specialized tokens to structurally distinguish thinking trajectories from final answers. The seed data pool includes both reasoning and non-reasoning data, which we filter based on strict quality control rules and emphasize the diversity, complexity, and comprehensiveness of the post-training data. In our experiments, we use 2 million data instances and experimentally determine the ratio of reasoning to non-reasoning samples to be 3:1.
We use a two-stage progressive optimization strategy with 6 epochs for each stage. The global batch sizes for each stage are set to 64 and 32, respectively.
The learning rate followed a cosine decay schedule, starting with an initial learning rate and gradually decreasing to 10\% of its peak during training. In order to achieve a better balance between convergence speed and training stability of the model, different initial learning rates of $8 \times 10^{-6}$ and $3 \times 10^{-6}$ are used at each stage of training.

\subsection{Experimental Results}

\paragraph{Evaluation Benchmarks}
We evaluate models derived from the \modelname~framework against a suite of strong baseline LLMs, including GPT-4o, DeepSeek-R1, Hunyuan-T1, QwQ-32B, and Qwen3-32B/14B, across six reasoning-oriented benchmarks categorized into: math, code, factual and knowledge-intensive reasoning, and logic tasks. The evaluation setting is the same as that of Pangu Ultra~\cite{Yin2025PanguUP}.

\begin{table}[htbp]
    \centering
    \caption{Comparison of \modelname~and other representative models across reasoning-oriented benchmarks. }
    \resizebox{\linewidth}{!}{
    \small
    \begin{tabular}{@{}l *{7}{c} @{}}
    
    \toprule
    \multirow{2}{*}{\centering\textbf{Model}} & \multirow{2}{*}{\textbf{AIME 2024}} & \multirow{2}{*}{\textbf{MATH-500}} & \textbf{GPQA} & \textbf{LiveCode} & \multirow{2}{*}{\textbf{ArenaHard}}& \multirow{2}{*}{\textbf{MMLU-pro}} & \multirow{2}{*}{\textbf{Avg.}}\\
    &  &  & \textbf{Diamond} & \textbf{Bench} &  \\
     \midrule
    {GPT-4o-0513} & 9.3 & 74.6 & 49.9 & 32.9 & 80.4 & 72.6 & 53.3\\
    {Hunyuan-T1} & 79.8 & 96.2 & 69.3 & 64.9 & 91.9 & 87.2 & 81.6\\
    {DeepSeek-R1} & 79.8 & 97.3 & 71.5 &  65.9 & 92.3 &  84.0 & 81.8\\
    {QwQ-32B} &  79.5 & 98.0 & 65.6 & 63.4$^\dagger$ & 89.5 & 79.6$^\dagger$ & 79.2\\
    {Qwen3-32B} & 81.4 & 97.2 & 68.4 & 66.3$^\dagger$ & 93.8 & 78.6$^\dagger$ & 80.9\\
    {Qwen3-14B} & 79.3 & 96.8 & 64.0 & 63.9$^\dagger$ & 91.7  & 76.2$^\dagger$ & 78.6\\
    \midrule
    {Pangu-38B} & 80.0 & 96.2 & 71.2 & 70.2 & 93.3 & 80.8 & 82.0\\
    {\modelname-1.6x} & 75.9 & 95.6 & {72.2} & {69.9} & {95.3} & 80.9 & 81.6\\
    {\modelname-2.1x} & 75.7 & 96.0 & 72.1 & 66.9 & 95.0 & 80.8 & 81.1\\
    {\modelname-4.2x} & 73.1 & 94.0 & 71.7 & 64.3 & 94.0 & 80.2 & 79.6\\
    \bottomrule
    \multicolumn{7}{l}{$\dagger$ indicates results from our internal evaluation.}
    \end{tabular}
    }
    \label{tab:reasoning-eval}  
\end{table}

\paragraph{Reasoning Capabilities and Comparative Benchmark Analysis}
Our empirical evaluation demonstrates the effectiveness of the \modelname~framework in accelerating Large Language Models while maintaining high performance. We first establish the strong capabilities of our base model, Pangu-38B, and then present the comprehensive results for the pruned \modelname~variants derived using our novel \modelname~methodology. The Pangu-38B model serves as our high-performing baseline, achieving a commendable average score of 82.0 across the suite of reasoning-oriented benchmarks detailed in Table~\ref{tab:reasoning-eval}. This strong foundation highlights the quality of the model from which our \modelname~variants are derived. Applying the \modelname~framework, we generated a series of structuredly pruned models achieving different acceleration ratios. These \modelname~variants exhibit highly competitive, and often superior, performance when compared to other representative models of similar effective sizes. For instance, a \modelname~variant achieving approximately 1.6x acceleration obtains an average score of 81.6, outperforming Qwen3-32B's score of 80.9 and performing comparably to Hunyuan-T1 at 81.6 and DeepSeek-R1 at 81.8. Similarly, a more aggressively pruned variant delivering approximately 4.2x acceleration attains an average score of 79.6, surpassing the Qwen3-14B model's score of 78.6. Another \modelname~variant, yielding roughly 2.1x acceleration, further reinforces these findings with a strong average score of 81.1, showcasing robust performance. These results collectively underscore the ability of \modelname~to significantly reduce model parameters while preserving critical reasoning capabilities across diverse and challenging benchmarks.

\begin{figure}[htbp]
    \centering
    \includegraphics[width=0.8\textwidth]{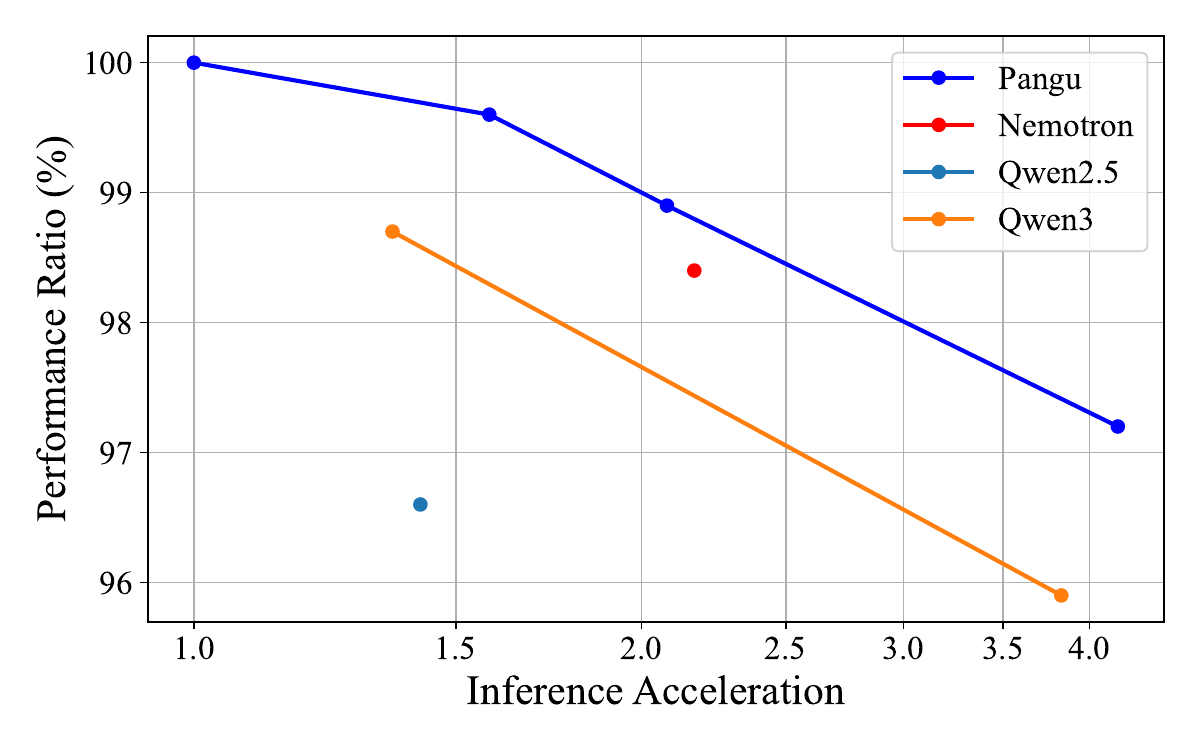}
    \caption{Performance ratio with respect to pruning ratio and acceleration ratio, illustrating the accuracy-efficiency trade-off. The \modelname~series exhibits a more favorable curve than those of both the Qwen3~\cite{qwen3} series and the PUZZLE~\cite{bercovich2024puzzle} framework.}
    \label{fig:pruning_curve}
\end{figure}

\paragraph{Superior Performance-Acceleration Trade-off}
The combination of sustained high accuracy on reasoning benchmarks and considerably increased throughput indicates that our \modelname~framework provides a superior performance-acceleration trade-off. This advantage is visually suggested by the pruning curve in Figure~\ref{fig:pruning_curve}, where the \modelname~series consistently operates at a more advantageous point on the performance versus acceleration/pruning ratio spectrum. To further contextualize this, we compare our results with recent work such as PUZZLE~\cite{bercovich2024puzzle}. Our \modelname-2.1x has an average score of 81.1 as detailed in Table~\ref{tab:reasoning-eval}, retains approximately 98.9\% of the original Pangu-38B's performance, which is higher than PUZZLE's reported 98.4\% retention with similar accleration ratio. This favorable comparison underscores the effectiveness of \modelname's principled component removal, sophisticated weight re-initialization techniques, and hardware-aware co-design in achieving significant LLM acceleration with minimal performance degradation.

\subsection{Ablation Study}
This section investigates the individual contributions of the key components proposed within our \modelname~pruning methodology, particularly our novel weight re-initialization strategies: Cross-Layer Attention Pruning (CLAP) and Stabilized LayerNorm Pruning (SLNP).

\begin{table}[htbp]
    \centering
    \caption{Ablation study results comparing pruning methods on various benchmarks. Avg scores are recalculated based on the six reported benchmarks: LAMBADA, WPLC, MMLU, C-Eval, BigBench, and HumanEval. CLAP denotes Cross-Layer Attention Pruning, and SLNP refers to Stabilized LayerNorm Pruning.}
    \resizebox{\linewidth}{!}{
    \small
    \begin{tabular}{lccccccc}
        \toprule
        \textbf{Method}& \textbf{LAMBADA} & \textbf{WPLC} & \textbf{MMLU} & \textbf{C-Eval} & \textbf{BigBench} & \textbf{HumanEval} & \textbf{Avg} \\
        \midrule
        Minitron~\cite{muralidharan2024compact}& 58.2 & 17.1 & 33.4 & 34.3 & 29.6 & 8.3  & 30.2 \\
        Minitron + CLAP  & \textbf{60.9} & 17.6 & 39.8 & 38.0 & 33.7 & 8.7& 33.1 \\
        Minitron + CLAP + SLNP  & 60.3 & \textbf{18.0} & \textbf{41.3} & \textbf{38.9} & \textbf{34.9} & \textbf{9.5} & \textbf{33.8}\\
        \bottomrule
    \end{tabular}
    }
    \label{tab:ablation}
\end{table}

\subsubsection{Impact of Weight Re-initialization Strategies}
To isolate and quantify the benefits of our proposed weight re-initialization techniques, we conduct an ablation study by incrementally adding them to a strong baseline pruning method, Minitron \cite{muralidharan2024compact}. These comparative experiments were performed using an experimental model pruned to 11B, which was subsequently fine-tuned on a dataset of 21 billion tokens for evaluation. The results, presented in Table~\ref{tab:ablation}, demonstrate the progressive improvements achieved by integrating CLAP and SLNP.

We then evaluate the addition of our Cross-Layer Attention Pruning and subsequently the combined effect of both Cross-Layer Attention Pruning and Stabilized LayerNorm Pruning, which represents our full weight re-initialization approach layered on the Minitron baseline.

\begin{table}[htbp]
    \centering
    \caption{Performance comparison of different post-normalization strategies on a 14B parameter model. }
    \resizebox{\linewidth}{!}{
    \small
    \begin{tabular}{lccccccc} 
        \toprule
        \textbf{Normalization Strategy} & \textbf{LAMBADA} & \textbf{WPLC} & \textbf{MMLU} & \textbf{C-Eval} & \textbf{BigBench} & \textbf{HumanEval} & \textbf{Avg} \\
        \midrule
        Sandwich                       & 76.4    & 34.2 & 73.7 & 75.2  & 61.7    & 38.1     & 59.9 \\
        Direct Prune                   & 72.6    & 28.2 & 64.6 & 62.0  & 51.0    & 29.0     & 51.2 \\
        DyT~\cite{zhu2025transformers} & 75.9    & 32.9 & 72.4 & 74.2  & 59.3    & 39.3     & 59.0 \\
        Norm Absorption             & 75.8    & 33.3 & 72.9 & 73.7  & 60.0    & 38.1     & 59.0 \\
        \bottomrule
    \end{tabular}
    }
    \label{tab:norm-ablation}
\end{table}
\paragraph{Analysis of Results}
The results presented in Table~\ref{tab:ablation} clearly demonstrate the efficacy of our proposed weight re-initialization strategies when applied over a competitive baseline. Adding Cross-Layer Attention Pruning (CLAP) to the Minitron baseline yields a significant improvement in the average score, from 30.2 to 33.1, an improvement of 2.9 points. This underscores the benefit of CLAP's mechanism for preserving and transferring salient attention information from layers designated for pruning, which is particularly beneficial for tasks requiring complex reasoning and knowledge retention, as suggested by the improvements on MMLU and C-Eval.

\begin{figure}[htbp]
    \centering
    \includegraphics[width=\textwidth]{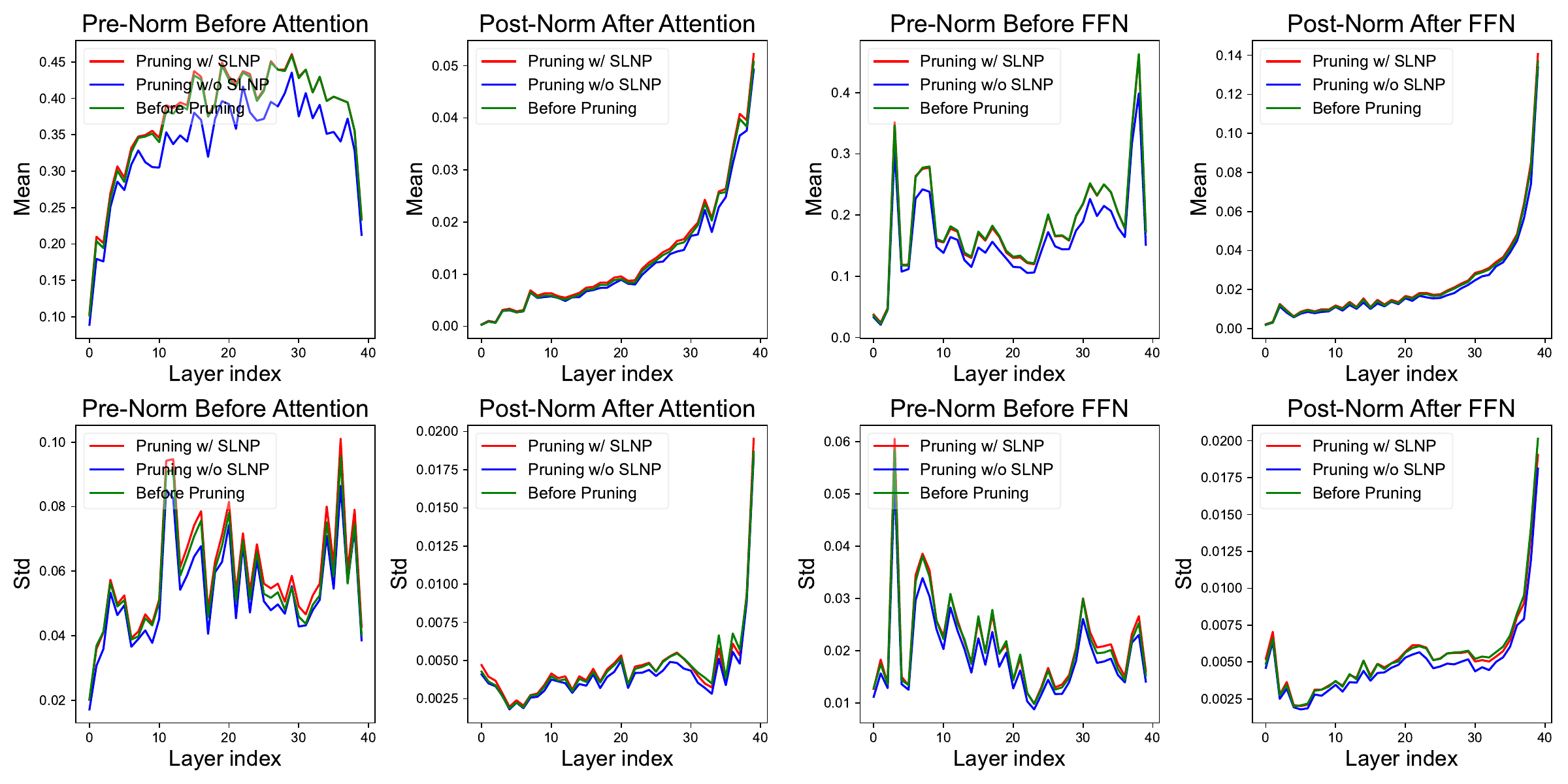} 
    \caption{Distribution of the Sandwich-Norm's affine scale parameters $\boldsymbol{\gamma}$ before and after pruning. Statistics, specifically mean and standard deviation, are shown for retained components within each relevant layer. The consistent distributions post-pruning, particularly after applying activation-based channel pruning with Stabilized LayerNorm Pruning (SLNP), indicate the stability of our method in preserving learned parameter characteristics.}
    \label{fig:pruning_norm}
\end{figure}

Subsequently, the application of Stabilized LayerNorm Pruning (SLNP) on top of the Baseline + CLAP configuration further elevates the average performance to 33.8. This is an increase of 0.7 points over the Baseline + CLAP configuration and a total increase of 3.6 points over the Baseline. This incremental gain highlights the importance of SLNP in stabilizing layer output statistics by re-initializing the affine parameters of RMSNorm layers, a crucial step when significant width pruning occurs concurrently with depth reduction. The consistent gains across multiple benchmarks, including the highest scores on MMLU, C-Eval, BigBench, and HumanEval for our full method, affirm that our weight re-initialization techniques are highly effective. They play a pivotal role in mitigating the substantial performance degradation typically associated with aggressive, joint width and depth pruning strategies, thereby enabling considerable model compression while maintaining stronger capabilities on downstream tasks.

\subsubsection{Comparison of Post-Normalization Strategies}
To mitigate the inference latency introduced by the Post-RMSNorm layers inherent in Sandwich-Norm architectures, we conducted a comparative study of several post-normalization replacement strategies. These experiments were performed on a convergent 14-billion parameter model, with performance evaluated after a fine-tuning phase using a dataset of about 9 billion training tokens. The results of this comparison are systematically presented in Table~\ref{tab:norm-ablation}.

The Sandwich configuration, which retains the complete Post-RMSNorm layers, serves as our performance upper bound, achieving an average score of 59.9 across the evaluated benchmarks. The experimental data clearly indicate that naively removing these Post-RMSNorm layers leads to a severe performance degradation, evidenced by a significant drop of 8.7 points in the average score, from 59.9 to 51.2. This substantial decline underscores the critical role these normalization layers play in maintaining model stability and performance, or alternatively, highlights the necessity of a well-designed mechanism if they are to be modified or removed for efficiency.

In stark contrast, our Post-RMSNorm absorption technique, detailed in Section~\ref{sec:sandwich_absorption_method} and referred to as the ``Norm Absorption'' method in Table~\ref{tab:norm-ablation}, demonstrates remarkable efficacy in preserving model performance. This approach, which transforms the dynamic Post-RMSNorm into a static, channel-wise scaling operation, achieves an average score of 59.0. This result indicates a substantial recovery of the performance lost through direct pruning, bringing the model's effectiveness remarkably close to that of the original, unoptimized Sandwich-Norm architecture, with the difference being only 0.9 average points.

Notably, our norm absorption method achieves performance on par with DyT~\cite{zhu2025transformers}, an alternative advanced technique that replaces Post-RMSNorm with a Dynamic Tanh activation. Both our method and DyT achieve an average score of 59.0. Both strategies effectively restore performance to a level nearly equivalent to the original Sandwich-Norm configuration. However, our norm absorption method offers this highly competitive outcome through a structurally simpler static scaling mechanism. As previously elaborated, this static scaling can be seamlessly fused into the weight matrices of adjacent linear layers. This fusion renders the original Post-RMSNorm operation effectively zero-cost during inference, presenting a compelling advantage in terms of both architectural simplicity and ultimate deployment efficiency while maintaining a high level of model accuracy.

\subsection{Parameter Analysis}
To assess the impact of our pruning methodology on the learned model parameters, particularly the affine scale parameters $\boldsymbol{\gamma}$ of the Sandwich-Norm layers, we analyze their distribution before and after pruning. This analysis focuses on the $\boldsymbol{\gamma}$ components corresponding to the channels and layers that are retained in the pruned model. For a fair comparison, when considering the ``before-pruning'' statistics, we specifically examine the original $\boldsymbol{\gamma}$ values associated with these subsequently retained structures. The ``after-pruning'' statistics reflect these same $\boldsymbol{\gamma}$ components following the application of our pruning techniques, including Stabilized LayerNorm Pruning (SLNP).

Figure~\ref{fig:pruning_norm} illustrates the distribution of these $\boldsymbol{\gamma}$ parameters, characterized by their mean and standard deviation, for each relevant normalization layer, both prior to and after the pruning and weight re-initialization process. As depicted, the mean and standard deviation of the $\boldsymbol{\gamma}$ values within each corresponding normalization layer remain remarkably consistent. This observation indicates that our activation-based pruning, coupled with the SLNP strategy, preserves the learned statistical properties of these critical scaling parameters, avoiding drastic distributional shifts. Such stability is desirable as it suggests that the core learned characteristics of the normalization layers are maintained, contributing to the overall robustness and predictable behavior of the pruned model.